\documentclass[lettersize,journal]{IEEEtran}
\usepackage{amsmath,amsfonts}
\usepackage{algorithmic}
\usepackage{algorithm}
\usepackage{array}
\usepackage{textcomp}
\usepackage{stfloats}
\usepackage{url}
\usepackage{verbatim}
\usepackage{graphicx}
\usepackage{cite}
\usepackage{CJKutf8}
\usepackage{amssymb} 
\usepackage{soul, color, xcolor}
\usepackage{tabularx,booktabs}
\usepackage{siunitx} 
\usepackage{subfigure}
\usepackage{multirow}
\usepackage{multicol}
\usepackage{threeparttable}
\usepackage[colorlinks,
            linkcolor=blue,
            anchorcolor=blue,
            citecolor=blue]{hyperref}
            
\begin{document}

\title{ElectricSight: 3D Hazard Monitoring for Power Lines Using Low-Cost Sensors}


\markboth{Journal of \LaTeX\ Class Files,~Vol.~14, No.~8, August~2021}%
{Shell \MakeLowercase{\textit{et al.}}: A Sample Article Using IEEEtran.cls for IEEE Journals}

\author{Xingchen Li$^{1}$, LiDian Wang$^{1}$, Yu Sheng$^{1}$, ZhiPeng Tang$^{1}$, Haojie Ren$^{1}$, Guoliang You$^{1}$, YiFan Duan$^{1}$,\\ Jianmin Ji$^{1,2}$, Yanyong Zhang$^{1,2}$

\thanks{$^1$ School of Computer Science and Technology, University of Science and Technology of China (USTC), Hefei 230026, China}
\thanks{$^2$ Institute of Artificial Intelligence, Hefei Comprehensive National Science Center, Hefei, Anhui, China}
\thanks{XingChen Li and LiDian Wang are co-first authors. }
\thanks{Corresponding author: Yanyong Zhang}
}

        


\maketitle

\begin{abstract}

Protecting power transmission lines from potential hazards involves critical tasks, one of which is the accurate measurement of distances between power lines and potential threats, such as large cranes.
The challenge with this task is that the current sensor-based methods face challenges in balancing accuracy and cost in distance measurement. A common practice is to install cameras on transmission towers, which, however, struggle to measure true 3D distances due to the lack of depth information. Although 3D lasers can provide accurate depth data, their high cost makes large-scale deployment impractical.

To address this challenge, we present ElectricSight, a system designed for 3D distance measurement and monitoring of potential hazards to power transmission lines. This work's key innovations lie in both the overall system framework and a monocular depth estimation method. 
Specifically, the system framework combines real-time images with environmental point cloud priors, enabling cost-effective and precise 3D distance measurements. As a core component of the system, the monocular depth estimation method enhances the performance by integrating 3D point cloud data into image-based estimates, improving both the accuracy and reliability of the system.

To assess ElectricSight's performance, we conducted tests with data from a real-world power transmission scenario. The experimental results demonstrate that ElectricSight achieves an average accuracy of 1.08 m for distance measurements and an early warning accuracy of 92\%.

\end{abstract}

\begin{IEEEkeywords}
3D distance measurement, Monocular depth estimation, Point cloud - image registration
\end{IEEEkeywords}

\section{Introduction}

\begin{figure}[t]
\centering
\includegraphics[width=0.45\textwidth]{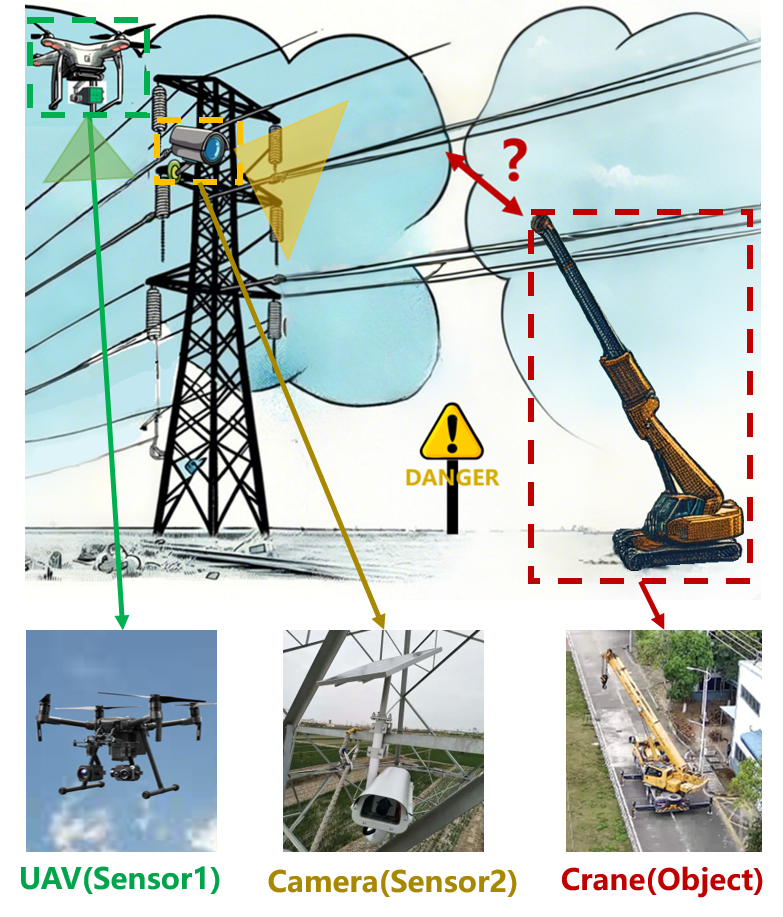}
\caption{Our distance measurement system: the UAV (sensor1) collects environmental point cloud data, while the fixed camera (sensor2) provides image information. The two work together to accurately measure the distance between hazardous objects (e.g., cranes) and power lines to detect potential risks.} 
\label{fig:intro}
\end{figure}

Power transmission is the critical component of the power system~\cite{raiola2018development, yoo2017mono, wang2019safety}, responsible for delivering electricity from the generation areas to various locations. Underground cables, though stable, are expensive and primarily used in Europe; overhead lines~\cite{zangl2009feasibility}, being less expensive, are widely used in China, the USA, and Japan. However, these exposed overhead lines are particularly vulnerable to threats from large objects like cranes and excavators, leading to potential widespread power outages. Therefore, ensuring the safety of transmission lines is an urgent issue to address.

Existing sensor technology can monitor the distance between physical threats and electrical wires and trigger alerts, significantly enhancing safety. However, commonly used 2D cameras~\cite{rw1} lack depth information, leading to insufficient monitoring accuracy. Although 3D laser sensors~\cite{rw18} are higher in performance, they are costly to deploy. Moreover, traditional manual inspections are time-consuming and labor-intensive, making it difficult to meet the efficient inspection demands of modern power grids. Therefore, developing an efficient, low-cost inspection solution is imperative.

We introduce a new power transmission line monitoring system named ElectricSight, which achieves high-precision, low-cost three-dimensional distance measurement by integrating image and environmental point cloud prior information. 
The system includes an innovative overall framework and a monocular depth estimation method that combines point cloud-image registration with geometric depth constraints. 

As shown in Fig.~\ref{fig:intro}, the system integrates cameras mounted on transmission towers and LiDAR carried by unmanned aerial vehicles (UAVs). The system uses the environmental point cloud data regularly collected by the UAVs to provide prior three-dimensional information, and constructs geometric depth constraints with the images captured by the cameras. This allows for accurate distance measurement and early warning of potential threats to the power lines. 
This design significantly enhances monitoring precision and efficiency. Additionally, it reduces system costs, as using a UAV to collect maps annually is much cheaper than installing numerous 3D sensors.



In the field of transmission line protection, combining image data with environmental point cloud data faces a key challenge: the point cloud data collected regularly by UAVs is not real-time and cannot include hazard objects that may appear at any moment, which are usually only visible in images. The core issue is how to integrate the non-real-time prior information from the point clouds with the real-time detection results from images for precise measurement of hazard objects. 

To address this challenge, we propose a method that combines existing point cloud-image registration technology with the construction of geometric depth constraints. This approach establishes a reliable 3D-2D correspondence, and based on this, it creates depth constraints that enable precise calculation of the shortest distances to power transmission lines.
By integrating point cloud and image data, our method significantly enhances the accuracy of distance measurements for potential hazards.

Experimental results show that the ElectricSight system achieves high accuracy in measuring 3D distances and monitoring potential threats, significantly outperforming existing methods. We validated the system using extensive data collected from diverse real-world scenarios, covering different hazardous objects, distances, and environments. These findings demonstrate its robustness, adaptability, and strong potential for enhancing the security of electrical systems.

In summary, the main contributions of this paper are as follows: 
\begin{enumerate}  
    \item We introduce ElectricSight, a 3D distance measurement system that combines image-based processing with prior knowledge from environmental point cloud maps. Specifically designed for power transmission line monitoring, it offers advantages in both high precision and low cost.   
    \item We propose a method that integrates high-precision point cloud-image registration with monocular depth estimation, utilizing geometric constraints. By effectively merging point cloud and image data, this approach significantly enhances the accuracy of hazard distance measurements and provides robust technical support for the system.
    \item Experimental results demonstrate ElectricSight's reliable and accurate performance across various scenarios. This includes different environments, multiple types of hazard objects, and a range of distances. The system achieves an average distance measurement error of approximately 1.08 m and reaches a 92\% accuracy rate in early warnings.
\end{enumerate}

\section{Related works}
\begin{figure*}[t]
\centering
\includegraphics[width=\textwidth]{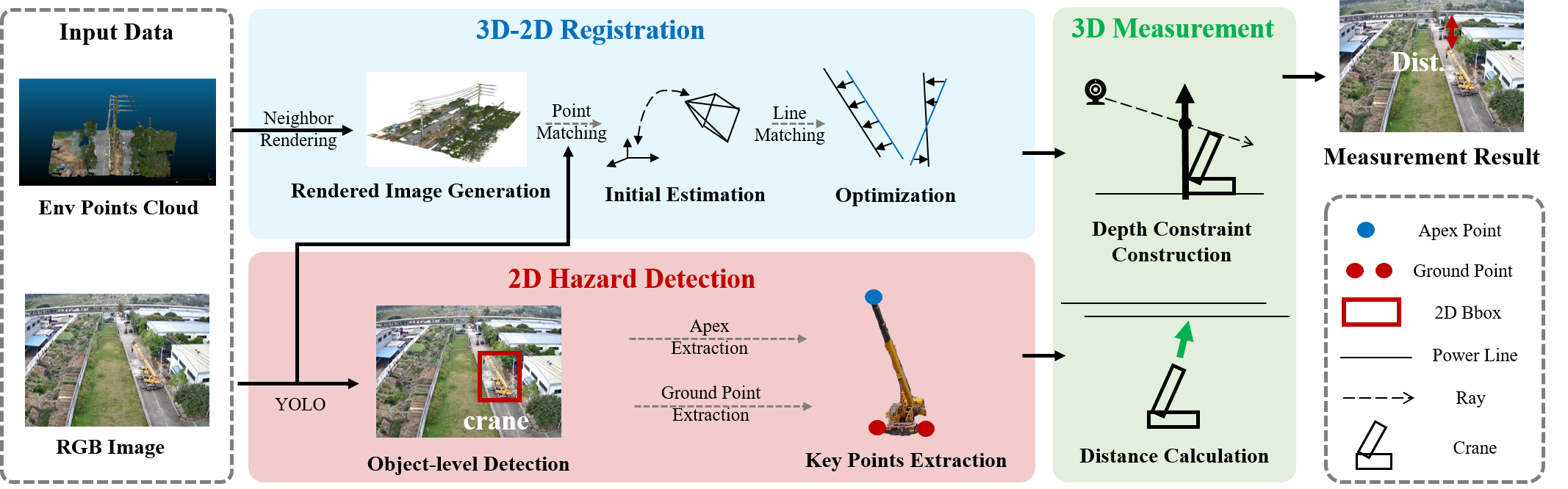}
\caption{Overview of our Electricsight: The method consists of three main stages: (1) 2D Detection: Detect hazardous objects (e.g., cranes) in 2D images and extract key points. (2) 3D-2D Registration: Perform initial estimation and optimization of the extrinsic parameters between point clouds and images using a neighbor rendering-based approach. (3) 3D Measurement: Construct depth constraints to accurately calculate the distance between hazardous objects and power lines, ensuring reliable safety assessment.}
\label{fig:pipeline}
\end{figure*}

Based on the type of sensors used, current automation threat detection in power systems can primarily be divided into two methods: one is based on two-dimensional (2D) sensor arrays, and the other is based on three-dimensional (3D) sensor arrays.

\subsection{2D-based Methods}

Methods based on 2D sensors, such as monocular cameras~\cite{gao2020linespyx, rw1}, thermal sensors~\cite{rw24}, and related devices, use 2D data for monitoring and detection. The widespread application of these methods is attributed to the simplicity and cost-effectiveness of the equipment. Research primarily focuses on the detection of power lines and hazardous objects, such as cranes, using image processing and machine learning algorithms. This is crucial for ensuring power grid safety and reliability.

As for transmission line detection, there has been considerable progress in this field. The PLGAN~\cite{rw1} utilizes generative adversarial networks to segment transmission lines from aerial images captured by unmanned aerial vehicles. 
Additionally, ~\cite{rw2} explores the use of stereo vision images from UAVs to identify transmission lines using an end-to-end convolutional neural network.
Recent advancements in deep learning have led to the development of 2D detectors for object detection~\cite{talukder2003real, cho2014multi}, significantly improving hazard detection. These advancements enable highly effective performance with minimal fine-tuning, even with limited data.

Despite challenges posed by the lack of 3D data, one study~\cite{rw11} successfully combines the detection of hazardous objects with distance measurement to high-voltage transmission lines. They use a monocular camera to reconstruct high-resolution images based on a sparse regularization model. They employ a sampling filter algorithm to extract transmission lines and construct a 3D model that automatically monitors the distances between high-voltage transmission lines and obstacles such as trees.

2D sensor technologies are favored for their low deployment costs and real-time monitoring capabilities, but the lack of real depth information remains a significant challenge for accurate distance measurement.

\subsection{3D-based Methods}
3D sensor-based methods utilize devices such as stereo cameras~\cite{rw17,rw25}, single-line lasers, and LiDAR sensors~\cite{rw13,rw18,rw19}. These technologies are characterized by their capability to capture depth information, enabling more precise spatial analysis and object recognition in 3D environments.

Stereoscopic vision provides enhanced 3D perception compared to monocular vision.~\cite{rw17} employs a stereo matching algorithm based on binocular vision to determine 3D coordinates of transmission lines and tree contours, calculating distances to obstacles. However, these measurements become unreliable at long ranges due to calibration errors. To overcome this limitation, integrating LiDAR sensors, which provide precise 3D data, is a viable solution.

LiDAR point clouds are used for 3D modeling of transmission lines~\cite{rw15, azevedo2019real, bian2019novel}. 
The study in ~\cite{rw15} enhances transmission line reconstruction by analyzing the distribution characteristics of power line groups and the contextual information of neighboring power tower objects. 
The research in \cite{rw13} uses a 3D catenary line model to iteratively fit segments of a power line point cloud, forming a continuous mathematical representation. This enables the direct calculation of the shortest distances between transmission lines and external objects, assessing their safety clearances. 
The study in \cite{rw18} uses LiDAR data from a drone, combining tree growth models and a two-stage intrusion detection algorithm to accurately detect the specific locations of tree intrusions.

However, the drawbacks of these methods include the high time and financial costs required to collect point cloud data and the challenges of achieving real-time monitoring. These factors impede the widespread applicability of such approaches.

\section{Method}
In this letter, we introduce a 3D distance measurement system.
This system primarily relies on images to identify potential hazards in power transmission lines, with environmental point clouds serving as auxiliary prior information, as shown in Fig.~\ref{fig:pipeline}. 
Section~\ref{sec:detection} first presents the hazardous object detection step in 2D images. 
This is followed by Section~\ref{sec:registration}, which details the registration of 2D images with 3D point clouds. Section~\ref{sec:measurement} details how we measure distances using depth constraints in 3D space.

\subsection{2D Hazard Detection}
\label{sec:detection}


First, to detect hazardous objects in 2D images, we implement a two-step process: 1) Object-level detection to identify the 2D bounding box of the object, and 2) Keypoint extraction to ascertain critical points, such as the apex (the highest point of the object) and the ground point (the object's contact with the ground). These keypoints are essential for distance measurement calculations in subsequent processes.

\subsubsection{Object-level Detection}
In the field of power transmission line protection, the types of potential hazards differ significantly from the typical objects found in traditional object detection datasets. For example, large objects such as cranes, excavators, and aerial lifts, if located near power transmission lines, can pose serious safety hazards. In contrast, typical objects such as cars or pedestrians pose almost negligible threats to power transmission lines.

To address this issue, we specifically construct a dataset for detecting hazardous objects on transmission lines. This dataset comprises images from various power transmission environments, featuring three high-risk hazardous objects: cranes, excavators, and aerial lifts. We manually annotate these objects and their extended arm components with 2D bounding boxes. For detailed information about the dataset, please refer to Section~\ref{sub:data}.

We implement the YOLOv7 model~\cite{wang2023yolov7} for object-level detection of potential hazards. We fine-tune YOLOv7, pretrained on the MS-COCO dataset, with a specially collected dataset of transmission hazard objects to tailor it more effectively to transmission scene detection tasks.

\subsubsection{Key Points Extraction}
From the object-level detection results (2D bounding boxes), we extract two types of key points: apexes and ground points. These key points are essential for the subsequent steps in distance measurement. In this context, key points refer to specific locations on an object, such as the apex (the highest point) and the ground point (the point of contact with the ground).

\paragraph{Apex Extraction}
\label{sub:apex}
The apex is the highest point of a potential hazard in the air, and its distance from the power transmission line usually reflects the shortest distance from the hazard object to the transmission line. Based on the structural characteristics of the hazard object, we use different strategies to extract the apex.

\textit{Aerial lift:} Because the working arm of the aerial lift always remains vertical, its apex is located at the center of the top of the bounding box. 
\textit{Crane or Excavator:} The booms of cranes and excavators have variable angle features and are inclined during operation. 
Initially, we apply the Gaussian Mixture Model (GMM)~\cite{stauffer1999adaptive} to segment the boom from the background image. Subsequently, we use the Hough Transform to detect its linear features. Through statistical voting, we determine the boom's inclination and select the most suitable top corner of the bounding box as the highest point.

\paragraph{Ground Point Extraction}
The ground point is defined as the point where the object touches the ground, crucial for assessing the object's depth. Assuming that the ground is flat, we use the midpoint of the bottom edge of the object's bounding box as the ground point.

Following the object detection and key point extraction processes, we identified N targets. Each target includes a vertex (Pt) and a ground contact point (Pg).

\subsection{3D- 2D Registration}
\label{sec:registration}
Obtaining 2D hazard detection results solely from images is insufficient; we also need to determine their positions in 3D space. 3D positional information is aided by environmental point cloud maps collected by UAVs.

To establish the correspondence between the environmental point cloud and camera images, we adopt the automatic registration method proposed by~\cite{sheng2024rendering}. This method renders the point cloud into photorealistic grayscale images from specific viewpoints. The rendered images reduce the differences in dimensions and appearance between the point cloud and camera images, significantly lowering the difficulty of registration while preserving the correspondence between 3D points in the point cloud and 2D pixel points.

According to the description in ~\cite{sheng2024rendering}, the automatic registration method consists of three parts:

\subsubsection{Neighbor Rendering} 
High-quality grayscale images are generated using specific attributes in the point cloud, such as RGB or intensity values. A Z-buffer is employed to filter occluded points.

\subsubsection{Initial Parameter Estimation} 
Based on the geometric prior assumption, initial camera poses are estimated to generate rendered images from multiple viewpoints. Subsequently, the feature matching algorithm SuperGlue is used to match feature points, and the rough extrinsic parameters are estimated using the EPnP algorithm.

\subsubsection{Parameter Optimization} 
The reprojection error is calculated by matching the point cloud with the image edge features. Optimization of the extrinsic parameters aims to minimize the sum of squares of these errors. To achieve this, the parameterization method of Lie groups and Lie algebras is used to map the optimization problem onto the SE(3) space, thereby reducing constraints and enhancing computational efficiency.

\subsection{3D Measurement}
\label{sec:measurement}

\begin{figure}[t]
\centering
\includegraphics[width=0.5 \textwidth]{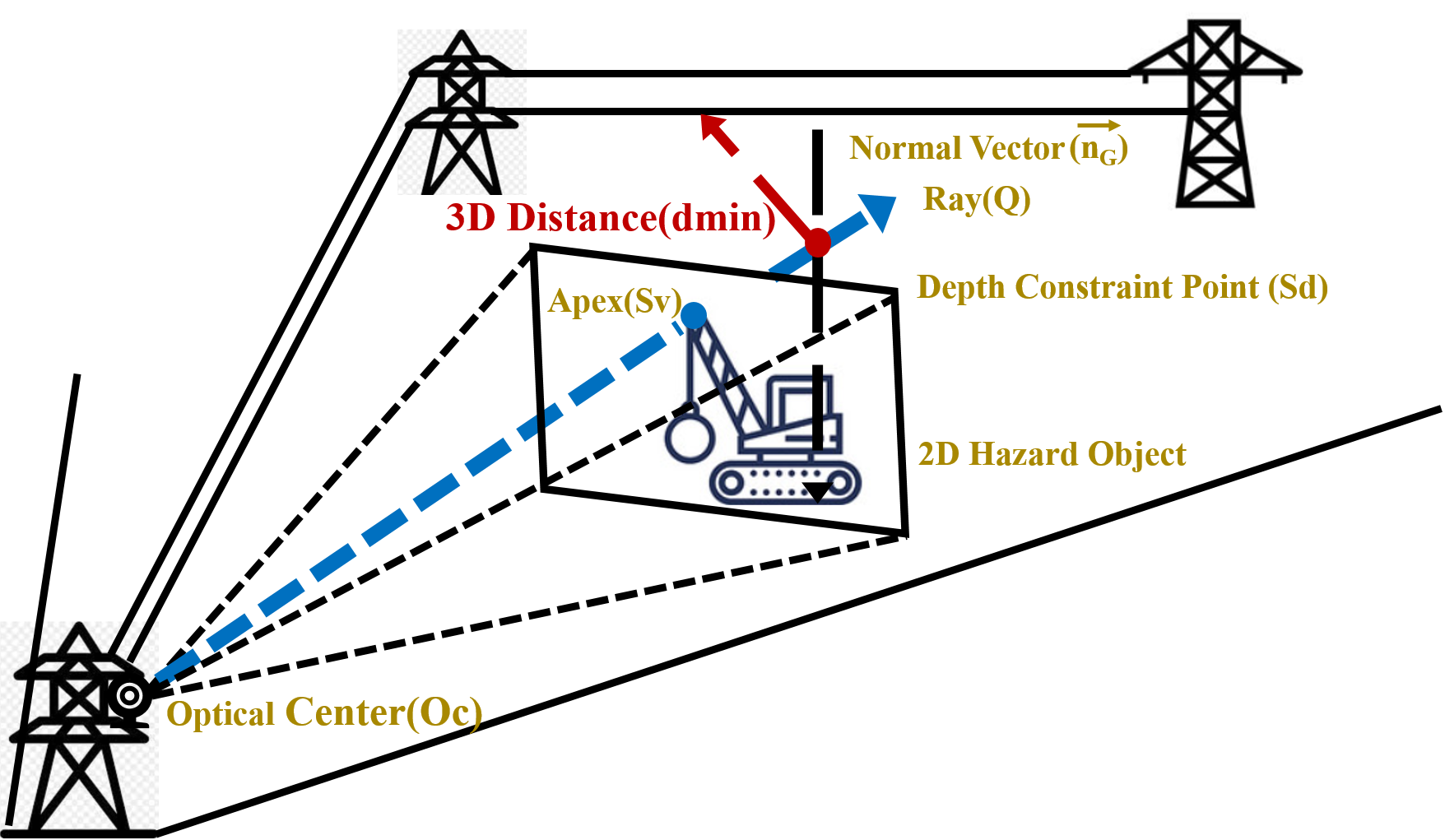}
\caption{Key steps of the 3D measurement stage, including ray construction, ground normal vector generation, depth constrain point creation, and final distance calculation.}
\label{fig:measure}
\end{figure}


In the 2D hazard object detection phase, we first obtain the bounding box of the object and identify its key points. 
These key points are then utilized in the 3D-2D registration process, where we calculate a transformation matrix $T_C^L$ to map the point cloud coordinates to the camera coordinates.
This step establishes the correspondence between 3D points and 2D pixels in the image.
In the 3D measurement stage, we combine the results of the first two steps and use geometric methods to calculate the shortest distance in 3D space from the hazard object to the power transmission line. Finally, we decide whether to issue a danger alert based on a predefined safety threshold.

The traditional 2D method relies on cameras to estimate the depth of objects~\cite{mobileye}, but it requires a large amount of prior knowledge. This method initially assumes that the ground is flat, with no slopes. Secondly, it requires the precise calculation of the vertical distance from the camera to the ground and the determination of the vanishing point~\cite{vanish}. 
However, applying these assumptions consistently and calculating the vanishing point in environments with complex visual obstructions, such as rugged terrains, utility poles, and vegetation, prove to be very challenging. Additionally, installing 3D sensors like laser rangefinders on power towers would significantly escalate costs.

To address the aforementioned issue, we utilize UAV-collected offline environmental point cloud maps and camera images for 3D distance measurement. This method strikes a balance between system accuracy, robustness, and cost. However, since the offline environmental maps does not provide 3D location information for dynamic hazard targets, depth estimation remains a critical component.

For regular-shaped objects such as aerial lifts, it is possible to approximate them using cuboids, which makes it easier to determine their position and orientation. However, for objects like cranes and excavators, which have irregular shapes due to long booms, it is challenging to estimate depth using geometric methods. To address this issue, we represent the entire object by key points, which allows for estimating the minimum distance \( l_{min} \) between key point \( S \) and power lines. Geometric constraints are added to ensure that the results closely mirror the actual ground conditions. The illustration of the entire process is shown in Fig.~\ref{fig:measure}. Specifically, the process is divided into the following steps:

\subsubsection{Depth Constraint Construction Based on Environmental Point Clouds}
\label{sub:depth con}
In the first step, we use Equation \ref{2D-3D trans} to project the offline collected environmental point cloud onto the image plane. 
For each pixel $S_C$ in the image, a corresponding 3D point $S_W$ can be established, generating a 2D-3D correspondence table $M_C^L$. This table is used to retrieve depth information for 2D pixels other than those obscured by real-time hazards. Subsequently, we utilize the depth data to provide constraints for the depth estimation of dynamic objects.

\begin{equation}
z_t
\begin{bmatrix}
S_C^T\\1
\end{bmatrix}
=
\begin{bmatrix}
    K & \Vec{0}^T
\end{bmatrix} T_C^L 
\begin{bmatrix}
    S_W^T\\1
\end{bmatrix}
\label{2D-3D trans}
\end{equation}

$S_W=(x,y,z)$ represents the coordinates of point $S$ in three-dimensional space, $S_C=(u,v)$ represents the pixel coordinates of point $S$ in the image, $T_C^L$ represents the extrinsic matrix from the point cloud coordinate system to the camera coordinate system, $K$ represents the intrinsic matrix of the camera, and $z_t$ is the depth scaling factor.

\subsubsection{Construction of the Ray Equation} In the second step, we identify a key point, the apex, from the results of 2D hidden object detection, denoted as $S_V = (u_v, v_v)$. Based on $S_V$, we construct the equation of the ray. The equation of the ray is as follows:

\begin{equation}
\begin{bmatrix}
    S_W \\1
\end{bmatrix}=
\frac{Orient
}{
||Orient||
}
t + 
\begin{bmatrix}
   O_C \\ 1 
\end{bmatrix},
t > 0
\end{equation}

where,
\begin{equation}
Orient=(T_C^L)^{-1}
\begin{bmatrix}
    [K^{-1} [S_V\ 1]^T]^T & 0
\end{bmatrix}^T
\end{equation}

Here, $O_C$ represents the position of the camera's optical center in the point cloud coordinate system, serving as the starting point of the ray, $Orient$ is the direction vector of the ray, pointing towards the keypoint, $S_W$ represents a point on the ray within the 3D point cloud coordinate system.

\subsubsection{Construction of the Ground Equation} In the third step, we first retrieve the coordinates of two grounding points among the keypoints from the 2D-3D correspondence table $M_C^L$, denoted as $A = (x_1, y_1, z_1)$ and $B = (x_2, y_2, z_2)$ in the 3D point cloud. Next, using the RANSAC algorithm, we estimate the ground plane and obtain the normal vector $\vec{n_G}$ of the ground. Based on $A$, $B$, and $\vec{n_G}$, we derive the equation of the plane $D$ that includes points $A$ and $B$: $$ ((B-A)\times\Vec{n_G})\cdot (X-A)=0 $$ Here, $X$ represents any point on the plane $D$.

\subsubsection{Based on Ray's Depth Constraint Construction} In the fourth step, we determine the depth constraint point $S_d$ by solving for the intersection of the plane equation $D$ and the ray $Q$. Subsequently, we update the apex of the ray to $S_d$, creating the modified ray $Q'$, described by the following equation:

\begin{equation}
\begin{bmatrix}
    S_w \\1
\end{bmatrix}=
\frac{Orient
}{
||Orient||
}
t + 
\begin{bmatrix}
   S_d \\ 1 
\end{bmatrix},
t > 0
\end{equation}

\subsubsection{Minimum Distance Calculation} In the final step, we calculate the minimum distance between the power line points and the hazard object. 
Given the difficulty in directly determining the 3D vertices of the hazard object in the air, we assume for our calculations that these apexes lie along a specific ray. 
We compute the shortest distance $d_{min}$ between the segmented power line points and the ray $Q'$. Finally, we compare $d_{min}$ with the predefined safety threshold $d_{thres}$. If $d_{min}$ falls below the threshold, we trigger a danger alert to ensure safety.

\section{Experiment}

\subsection{Data Preparation}
\label{sub:data}
\subsubsection{Low-cost Hardware Systems and Data Acquisition}
As shown in Fig.\ref{fig:intro}, we have constructed a low-cost hardware system for measuring the distance of potential hazard objects in real transmission corridors using offline environmental point clouds and real-time monocular images. Specifically, we have mounted a high-resolution monocular camera on the transmission tower to capture images, and separately, deployed a Matrice 300 RTK UAV equipped with a DJI L1 sensor platform (featuring two RGB cameras and a Livox LiDAR) for constructing environmental point cloud maps and collecting ground truth data, including hazardous objects. Additionally, we have utilized various common engineering vehicles, such as cranes, excavators, and aerial lifts, to serve as target hazards. The entire data acquisition process was completed in a secure experimental base.

On this hardware system, we conduct the following data collection steps:
\begin{itemize}
    \item \textit{Point cloud map acquisition:} The UAV conduct extensive mapping of the experimental site. It generate high-quality color point clouds at a density of 400 points per square meter. These point clouds are crucial for registering with monocular images and providing depth information for subsequent 3D measurement processes.
    
    \item \textit{Engineering operation simulation:} Professional operators operate engineering vehicles to simulate construction operations at different locations within the transmission corridor, preparing for various potential hazard scenarios that may be encountered in real-world environments. The operations involve deploying supports, rotating the structure, lifting and extending booms. These actions prepare for handling hazardous situations during construction at different locations, angles, and boom lengths.
    
    \item \textit{Monocular image acquisition:} We utilize a 4G network for real-time data transmission from the high-definition camera on the transmission tower, capturing images at a rate of 1Hz for detecting and measuring hazardous objects.
    \item \textit{Ground truth point cloud acquisition:} For each simulated construction state of the engineering vehicle, we utilize a UAV to collect point clouds from the area. This results in a colored point cloud map that encompasses both the current environment and the engineering vehicle, thus providing ground truth data for 3D measurements.
\end{itemize}

\subsubsection{Ground Truth Annotation and Evaluation Criteria}
To assess performance across different components, we collect various ground truth data and use specific evaluation criteria: 
\begin{itemize}
    \item \textit{2D Hazard Detection:} We annotate 6518 images with a total of 12035 bounding boxes. 80\% of the data is used as the training set, with 10\% each for validation and testing. The performance of 2D Hazard Detection is evaluated using mean Average Precision (mAP).
    \item \textit{3D-2D Registration:} 
    According to~\cite{koide2023general}, we uses survey-grade high-reflectance cones to estimate the transformation parameters between the UAV LiDAR and the camera. For each set of point cloud-image data, we annotate 50 pairs of 3D-2D matching points. These annotations help identify the LiDAR-camera transformation parameters that minimize the reprojection error of the target. We annotate a total of 90 pairs of point cloud-image data and use the estimated transformation as the "pseudo" ground truth.
    \item \textit{3D Measurement:} For 90 ground truth point clouds containing hazardous objects, we identifiy the 3D point on each hazard that was closest to the power line, calculating the shortest distance from this point to the wires' point cloud to serve as the ground truth for the algorithm. We use the average error (referred to as Distance Error) between the estimated shortest distance and the ground truth as the performance evaluation metric for the entire system. What's more, we have also incorporated the alarm accuracy to evaluate the performance of our system's warning function. The system issues an alert for hazards that are too close to the power lines based on a threshold. A correct alarm occurs when any part of a hazard is within the threshold distance from the wires, prompting the system to issue a danger alert. Here, we set the threshold to 10 meters.
\end{itemize}

\subsection{Evaluation of 3D Measurement}

\begin{table*}[tbp]
\begin{threeparttable}
    \caption{The results of 3D Measurement for Different Types of Target on our own dataset}
    \label{tab:comparison}
    \centering
    \begin{tabularx}{\textwidth}{X|X|X|XXX|X}
        \toprule
        \multirow{2}{*}{Method} & \multirow{2}{*}{Input Modality} & \multirow{2}{*}{Method Type} & \multicolumn{3}{c|}{$\text{Mean Distance Error}$$(m)\downarrow$} & \multirow{2}{*}{Alarm Accuracy$(\%)$} \\
         & & & crane-like & lifts-like & All &  \\
        \hline
        \midrule
        Metric3Dv2~\cite{hu2024metric3d} & Image Only & Learning & $4.221_{\pm11.060}$ & $3.989_{\pm1.528}$ & $4.121 _ { \pm 8.398}$ & 65.9\\
        Mobileye~\cite{1212895}  & Image Only & Geometry & $2.975_{\pm4.227}$ & $1.998_{\pm1.420}$ & $2.553_{\pm3.355}$ & 88.6 \\
        Ours & Image+Point cloud & Geometry & $\mathbf{1.496_{\pm1.255}}$ & $\mathbf{0.534_{\pm0.495}}$ & $\mathbf{1.081_{\pm1.108}}$ & $\mathbf{92.0}$\\
        \bottomrule
    \end{tabularx}
\end{threeparttable}
\end{table*}

\begin{figure}[t]
\centering
\includegraphics[width=0.5 \textwidth]{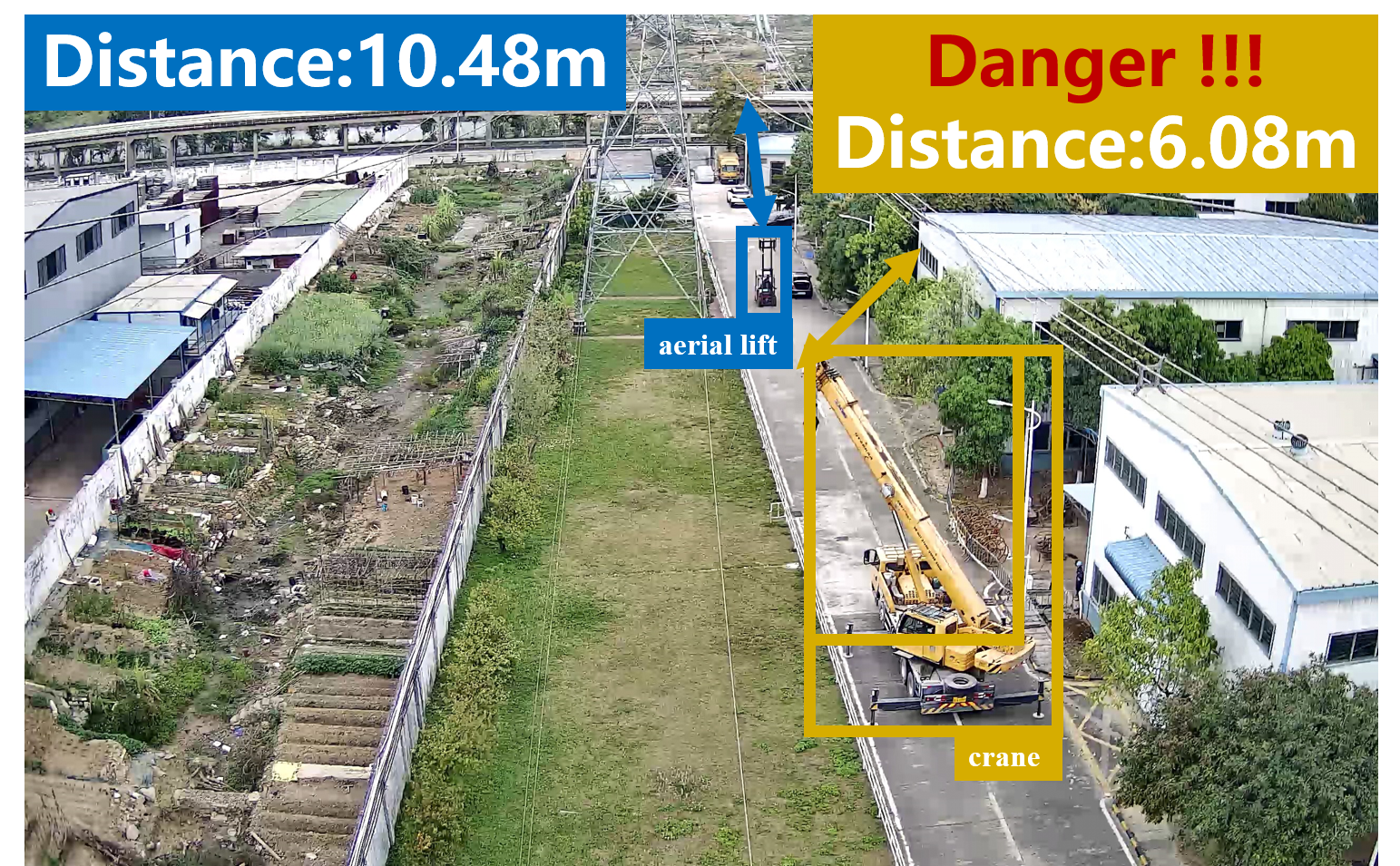}
\caption{The visualization of distance measurement and warning.
Crane: Distance = 6.08m (below 10m threshold), warning triggered.
Aerial lift: Distance = 10.48m (above 10m threshold), no warning.}
\label{fig:result}
\end{figure}

We evaluate the 3D measurement performance of our method by comparing the distance errors of two types of objects with the following methods:
(1) Metric3Dv2~\cite{hu2024metric3d}, which implements state-of-the-art zero-shot metric monocular depth estimation, was used in our study. For Metric3Dv2, we selected ViT-Large as the backbone and incorporated the camera's intrinsic parameters to enhance performance. 
(2) Mobileye~\cite{1212895} is a method that utilizes traditional computer vision techniques to estimate monocular depth through geometric approaches. This method leverages known target object height information as a prior constraint to estimate depth.
(3) Our ElectricSight.

Among these methods, Metric3Dv2 and Mobileye utilize only monocular images. In contrast, our approach incorporates an offline environmental point cloud map. Furthermore, while Metric3Dv2 employs a resource-intensive learning-based method, our approach and Mobileye use less resource-intensive geometric calculations.

Table~\ref{tab:comparison} summarizes the results on a 200-meter transmission corridor. In the results of all types of hazardous objects, we observe that our approach significantly surpasses the other methods. We also present a detailed analysis of the observations. The qualitative result is shown in Fig.~\ref{fig:result}.

From Table \ref{tab:comparison}, We observe that our ElectricSight significantly outperforms the other two methods with a mean distance error of 1.081 meters across all objects. 
Compared to Mobileye and Metric3Dv2, our method achieves a reduction of 58.1\% and 73.8\% in mean distance error, respectively, while also demonstrating greater stability with a standard deviation of 1.108. 
Additionally, in terms of alarm accuracy rates, ours system achieves 92.0\%, which is 3.4\% and 26.1\% higher than Mobileye and Metric3Dv2 respectively.

In our view, Metric3Dv2 and Mobileye, which only use image, struggle to estimate depth due to the lack of indirect depth information and conditional priors. 
Mobileye's approach precise target object height to accurately estimate depth. However, accurately determining the height of hazards, such as cranes, is inherently challenging. 
As for Metric3Dv2, it performs poorly in our scenario, failing to demonstrate its claimed zero-shot scene generalization capability. 
Conversely, Our ElectricSight obtains low-cost depth information by collecting offline point cloud maps, and since our solution is based on geometric methods, it does not require the high-cost training data collection which is necessary for learning-based schemes.

\subsection{Intermediate Results}
On the test set of hazardous object detection, our detection method achieves a mean Average Precision (mAP@[0.5:0.95]) of 96\%, demonstrating excellent performance in identifying hazardous objects in the specific context of power transmission lines. Furthermore, the method achieves a class-specific mAP of 98\% for cranes, 95\% for excavators, and 96\% for aerial lifts. 

As for 3D-2D registration, our registration method achieves an average translation error of 0.08m and a rotation error of $0.20 \, \si{\degree}$ across various electrical scenarios. This demonstrates that our method can effectively align 3D and 2D data.

\subsection{Ablation Studies}

\begin{table}[tbp]
\begin{threeparttable}
    \caption{Ablation Study on methods of Hazard Detection and 3D Measurement parts}
    \label{tab:ablation}
    \centering
    \begin{tabularx}{0.50\textwidth}{ccc|X}
        \toprule
        {Depth Constraint} & {$\text{Detection}_\text{arm}$} & {GMM \& HT} & {$\text{Mean Distance Error}$$(m)\downarrow$} 
        \\
        \hline
        \midrule
        \checkmark & & & $1.769_{\pm1.330}$ \\
        \checkmark & \checkmark & & $1.770_{\pm1.304}$ \\
        & \checkmark & \checkmark & $2.878_{\pm3.934}$\\
        \checkmark & \checkmark & \checkmark & $\mathbf{1.518_{\pm1.242}}$  \\
        \bottomrule
    \end{tabularx}
    \begin{tablenotes}[flushleft]
        \item \textbf{Depth Constraint} represents the use of environmental point clouds to obtain the depth range of the target, $\textbf{Detection}_\textbf{arm}$ represents the recognition of the crane arm, and \textbf{GMM \& HT} represents the method using Gaussian Mixture Model (GMM) and Hough Transform to assist in extracting the target vertices.
    \end{tablenotes}
\end{threeparttable}
\end{table}

In this section, we conduct ablation studies on the two key methods: Apex Extraction and Depth Constraint. Table~\ref{tab:ablation} summarizes the experimental results of different approaches. To draw significant conclusions, we conduct experiments solely on the crane category hazards, which have the highest demands for apex and depth accuracy.

\subsubsection{Apex Extraction}
We first investigate the impact of the apex extraction part discussed in Section~\ref{sub:apex}. 
For the bounding boxes of the booms, we compare the results using the midpoint of the top of the crane box with the midpoint of the top of the boom box as the apex. From the first two rows of Table~\ref{tab:ablation}, it can be seen that the results are almost identical, indicating that adding the boom box alone does not enhance performance.
However, our method's performance on crane category hazard data, compared to the two methods without using GMM and Hough Transform, shows an increase in Mean Distance Error of over 0.2m (see Table~\ref{tab:ablation}, comparing the 4th row with the 1st and 2nd rows).
This proves that the apex extracted through GMM and Hough Transform methods can significantly enhance the accuracy of our system.

\subsubsection{Depth Constraint}
Next, we discuss the impact of depth constraints, as detailed in Section~\ref{sub:depth con}. 
In our method, we calculate depth constraints based on the actual environment by detecting ground points and environmental point clouds.
Compared to the results obtained by using only geometric methods to calculate depth constraints (see Table~\ref{tab:ablation}, the 3rd row), our method can increase the Mean Distance Error by 47.3\% on the crane category. 
This indicates that using environmental point clouds as priors to construct depth constraints is crucial for enhancing the performance and stability of the system.

\subsection{Analysis}

\begin{figure}[t]
\begin{center}
	\subfigure[]{
	\begin{minipage}{0.45\linewidth}
        \centerline{\includegraphics[width=\textwidth]{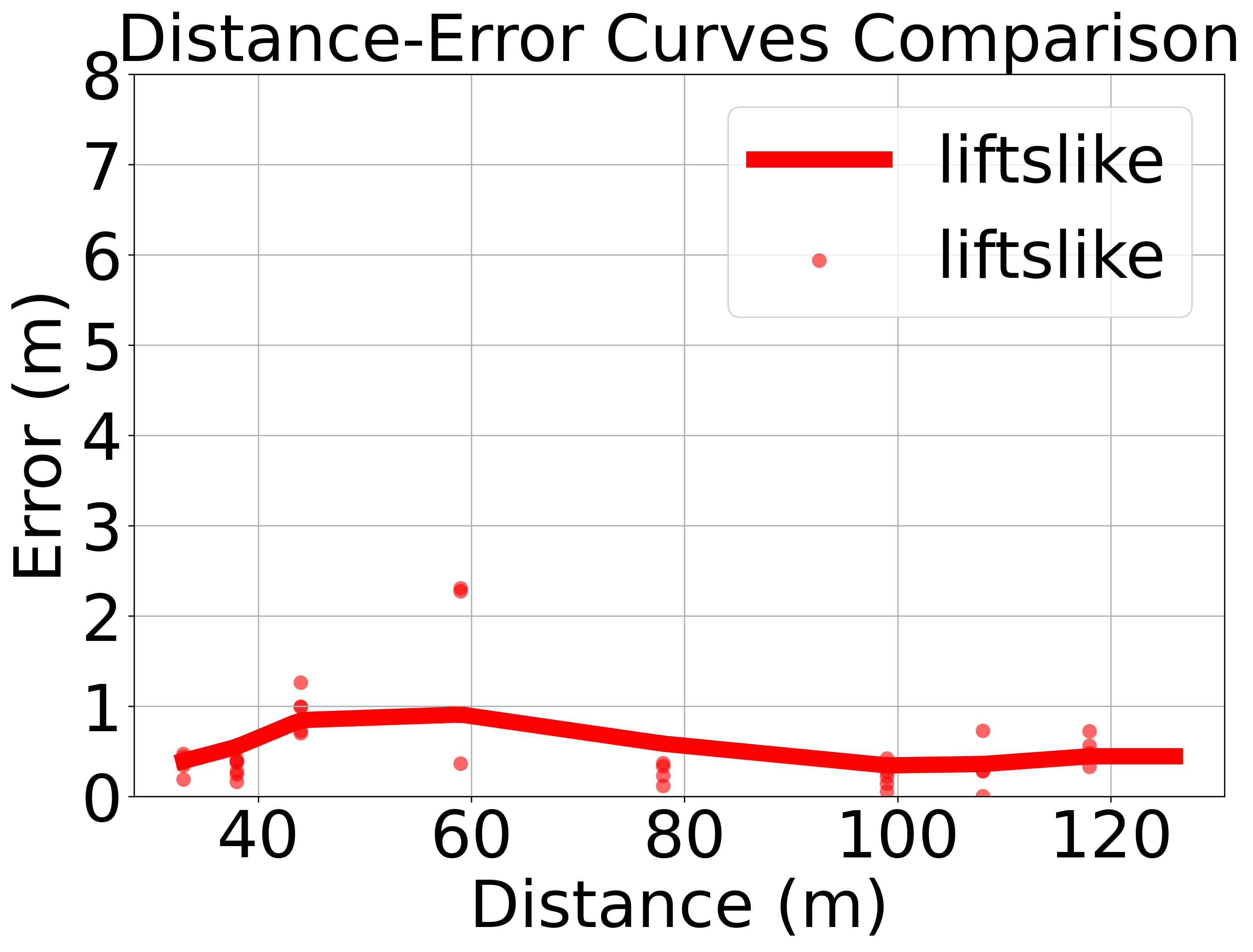}}
        \vspace{0.3em}
	\end{minipage}
        }
 	\subfigure[]{
	\begin{minipage}{0.45\linewidth}
        \centerline{\includegraphics[width=\textwidth]{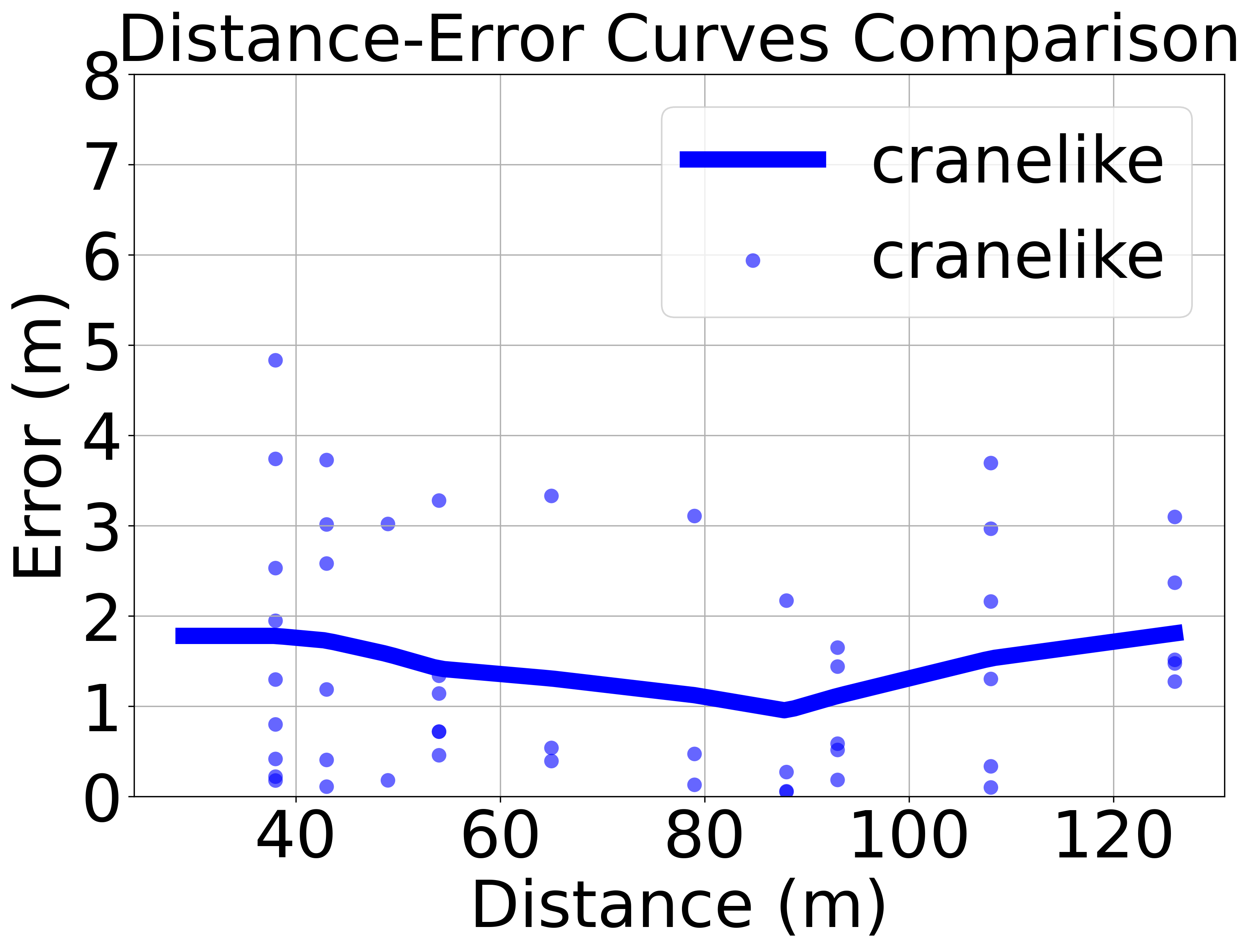}}
        \vspace{0.3em}
	\end{minipage}
        }
\caption{3D Measure results of different target types in 20-140m range: (a)\textit{Lifts-like} targets, (b)\textit{Crane-like} targets.}
\label{fig:analysis}
\end{center}
\vspace{-1.0em}
\end{figure}

The results of 3D Measurement can be affected by various factors. This section will analyze the impact of the type and distance of the target object on measurement accuracy.

\subsubsection{Type of Target Object}
Our 3D measurement method depends on accurately localizing target points in three dimensions. From Equation~\ref{2D-3D trans}, it is known that deriving the corresponding 3D coordinates from 2D points requires accurate real depth of the target point. The challenge of estimating the depth of the apex of different types of objects varies greatly. For objects like elevators, as shown by Fig.~\ref{fig:analysis}(a), our method achieves more stable and accurate results at all distances. In our view, the position of the top of elevator-type targets is distributed within a small range, and it's close to the target's ground point in the depth direction of the camera. Therefore, even when we use the depth of the ground contact point instead of the apex depth, we can obtain a small error. However, for crane-type targets, the apexes are distributed over a depth range exceeding the vehicle's length, resulting in larger errors when approximating their depth using the ground point. The Fig.~\ref{fig:analysis}(b)'s line and the distribution of the scatter points illustrate this point.

\subsubsection{Distance}
Mobileye~\cite{1212895} points out that the depth estimation error caused by pixel error changes exponentially with the real depth. The error formula is as follows: 

\begin{equation}
Z_{err}=Z_n-Z=\frac{fH}{y+n}-Z=\frac{nZ^2}{fH+nZ}
\label{eq:depth err}
\end{equation}

where $Z_n$ represents the depth calculated with $n$ pixel errors, $Z$ represents the real depth, $f$ represents the focal length, $y$ represents the pixel distance on the image, and $H$ represents the corresponding real-world distance. 
This formula indicates that the greater distance the estimated target is, the greater the resultant depth error caused by a single pixel. However, as shown in Fig.~\ref{fig:analysis}, we observe that results of our method are independent of distance. In our view, we approach this task by calculating the distance between a ray and a curve, not by estimating the 3D position of the target point, making it less reliant on depth accuracy.
Moreover, by using real environmental point clouds, we obtain a real depth constraint, which significantly reduces the errors caused by depth absence and thereby enhances the stability of our method over distance.

\begin{figure}[t]
\centering
\includegraphics[width=0.45 \textwidth]{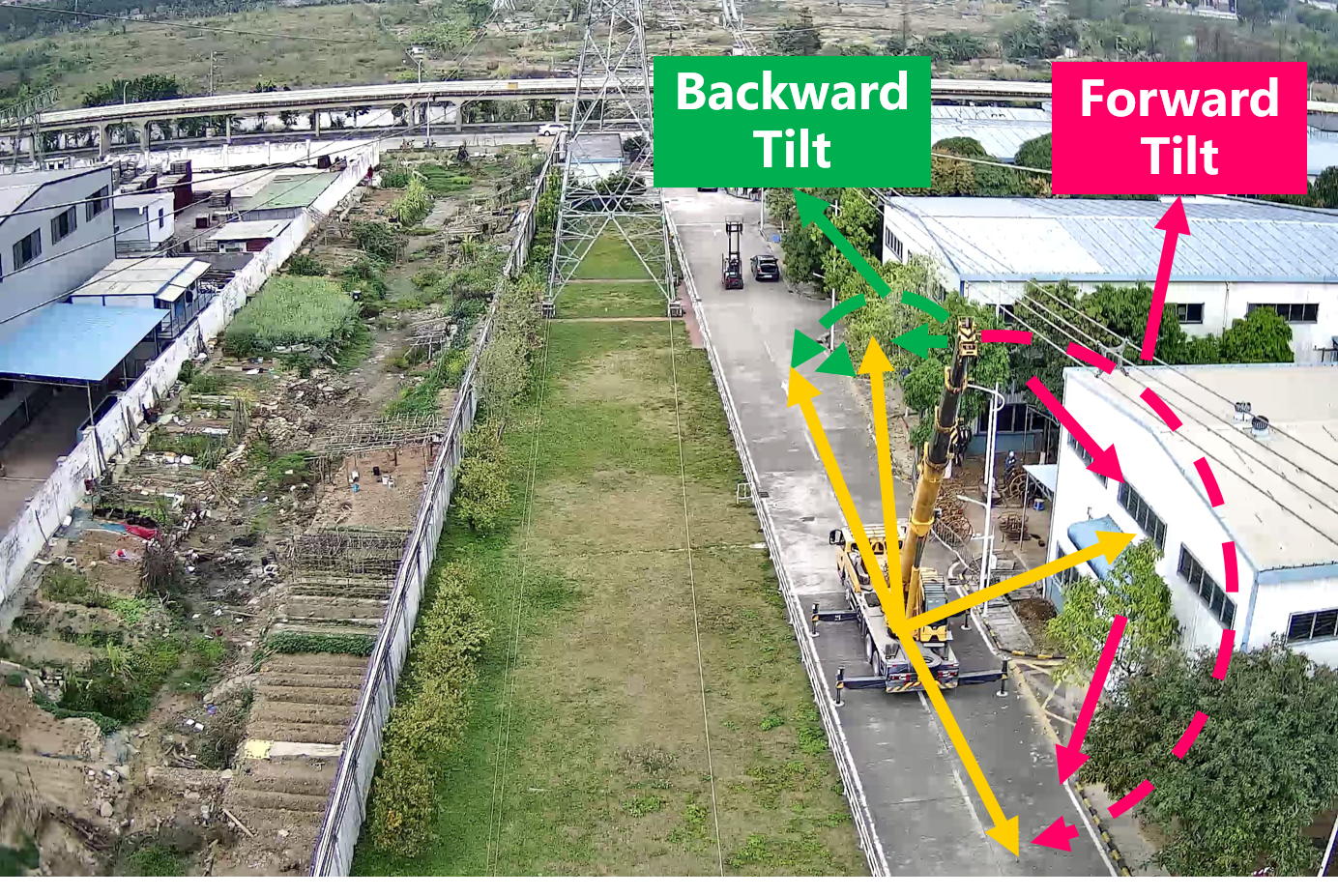}
\caption{Illustration of the crane boom's forward and backward tilt.}
\label{fig:badcase}
\end{figure}

\subsubsection{Bad Case}
Since we calculate the minimum distance between a ray and a curve instead of performing precise 3D position estimation of the target, there are some situations we cannot handle. 
As shown in Fig.~\ref{fig:badcase}, the crane boom is tilted forward, that is, inclined towards the direction of the camera. This causes the apex of the crane boom not to be located on the ray, and thus the actual depth exceeds the estimated range. Due to this discrepancy, the estimated distance error is significant, reaching 4.832 meters.

\section{Conclusion}
In this paper, we introduce ElectricSight, an innovative system designed to measure the distance between hazardous objects and transmission lines in power scenarios.
Our system relies on images and uses pre-acquired environmental point clouds as prior information. It is automated, accurate, and cost-effective.
In addition, we design an innovative method that combines the 3D prior obtained by point cloud-image registration with a monocular depth estimation algorithm based on geometric depth constraints.
This approach accurately measures the distance of potential hazards from 2D images.
Experiments in various scenarios show that our method has reliable and accurate performance for distance measurement in transmission line protection scenarios.

\bibliographystyle{IEEEtran}
\bibliography{refs}

\end{document}